\documentclass[runningheads]{llncs}
\usepackage{graphicx}
\usepackage[caption=false]{subfig}
\usepackage{hyperref}
\usepackage{cite}
\usepackage{wrapfig}
\hypersetup{hidelinks,
	backref=true,
	pagebackref=true,
	hyperindex=true,
	breaklinks=true,
	colorlinks=true,
	urlcolor=blue,
	bookmarks=true,
	bookmarksopen=false,
	pdftitle={Title},
	pdfauthor={Author}}
\begin{document}
\title{Self-Adapting Goals Allow Transfer of Predictive Models to New 
Tasks}
%
%
\author{Kai Olav Ellefsen\inst{1} \and
Jim Torresen\inst{2}}
\authorrunning{K.O. Ellefsen \and J. Torresen}
%
\institute{Department of Informatics, University of Oslo, Norway \and
Department of Informatics and RITMO, University of Oslo, Norway
\email{kaiolae@ifi.uio.no}}
\maketitle              
\begin{abstract}
A long-standing challenge in Reinforcement Learning is enabling agents to learn 
a model of their environment which can be transferred to solve other problems 
in a world with the same underlying rules. One reason this is difficult is the 
challenge 
of learning accurate models of an environment. If such a model is inaccurate, 
the agent's plans and actions will likely be sub-optimal, and likely lead to 
the wrong outcomes. Recent progress in model-based reinforcement learning has 
improved the ability for agents to learn and use predictive models. In this 
paper, we extend a recent deep learning architecture which learns a predictive 
model of the 
environment that aims to predict only the value of a few key measurements, 
which are indicative of an agent's performance. Predicting only a few 
measurements rather than the entire future state of an environment makes it 
more feasible to learn a valuable predictive model. We extend this predictive 
model with a small, evolving neural network that suggests the best goals to 
pursue in the current state. We demonstrate that this allows the predictive 
model to transfer to new scenarios where goals are different, and that the 
adaptive goals can even adjust agent behavior on-line, changing its strategy to 
fit the current context.

\end{abstract}
%
%
\section{Introduction}


Humans and animals rely on internal models (mental simulations of how objects 
respond to interaction) to predict the consequences of their actions and 
generate accurate motor commands in a wide range of 
situations~\cite{Schillaci2016, Wolpert2003}. Recent advances in deep 
learning have enabled computers to learn such predictive 
models by gathering a large collection of observations from an 
environment~\cite{Luc2017,2017arXiv170405831V}. This opens up the 
possibility for guiding the actions of robots and computer agents by having 
them predict the consequences of each action and selecting the one leading to 
the best outcome.

When using predictions for guiding the actions of an agent, it is beneficial to 
limit the prediction to the most essential parts of the environment. For 
instance, if we would like to predict the effect of turning the steering wheel 
of a car, we should not try to predict the effect this has on birds we see in 
the 
sky, trees far away from us, and so on. Rather, we should focus on the effect 
on 
some key observable measurements, such as the car speed, and the distance to 
other 
cars and pedestrians. If we focus on 
predicting the parts of the environment we currently care about, the prediction 
problem becomes much more manageable.

This intuition is the background for a recent, popular technique for learning 
to act by predicting the future~\cite{Dosovitskiy2017}. The technique makes the 
assumption that we can analyze the outcome of an action by focusing on \emph{a 
few measurements}. These measurements should be the observable quantities that 
are most related to success or failure in some scenario. The 
authors propose that a way to learn how to act in an environment is to learn 
to predict the effect one's actions have on these measurements. Such 
predictions 
can readily be learned by gathering a large set of examples of observations, 
actions and resulting measurements from the environment (e.g. by simulating 
thousands of car trips).

Once one has learned to predict the measurements resulting from one's actions, 
it is possible to select the best action for a given situation, by choosing the 
one giving the \emph{most optimal predicted measurements}. This requires some 
way to map the predicted future measurements to a number representing the 
\emph{utility} of this future. \cite{Dosovitskiy2017} solved this by defining 
a goal vector, which weights the different measurements according to 
user-defined rules. For instance, a rule could be that we give a very high 
weight to the measurement of distance to the nearest pedestrian, and a lower 
positive weight to the measurement of car speed, reflecting that we want to 
drive efficiently, but only if pedestrians around us are safe.

\begin{figure}
	\centering
	\subfloat[The Goal-ANN produces goals adapted to the current 
	situation]{\includegraphics[width=0.8\textwidth]{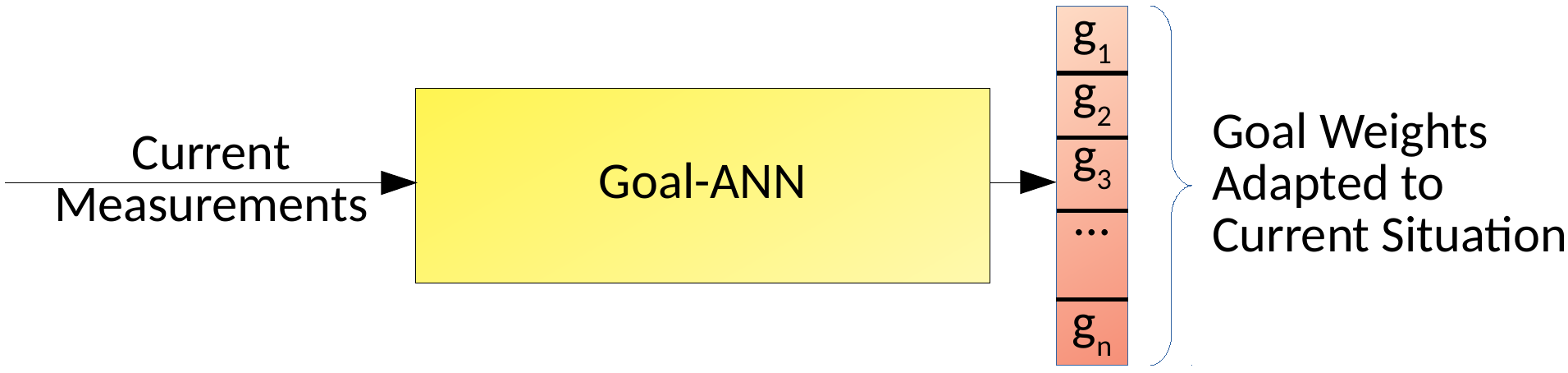}}
	
	\hspace{0.25cm}
	
	\subfloat[The predictive ANN (developed by~\cite{Dosovitskiy2017}) predicts 
	the 
	consequence of taking each action in the current situation.] 
	{\includegraphics[width=0.8\textwidth]{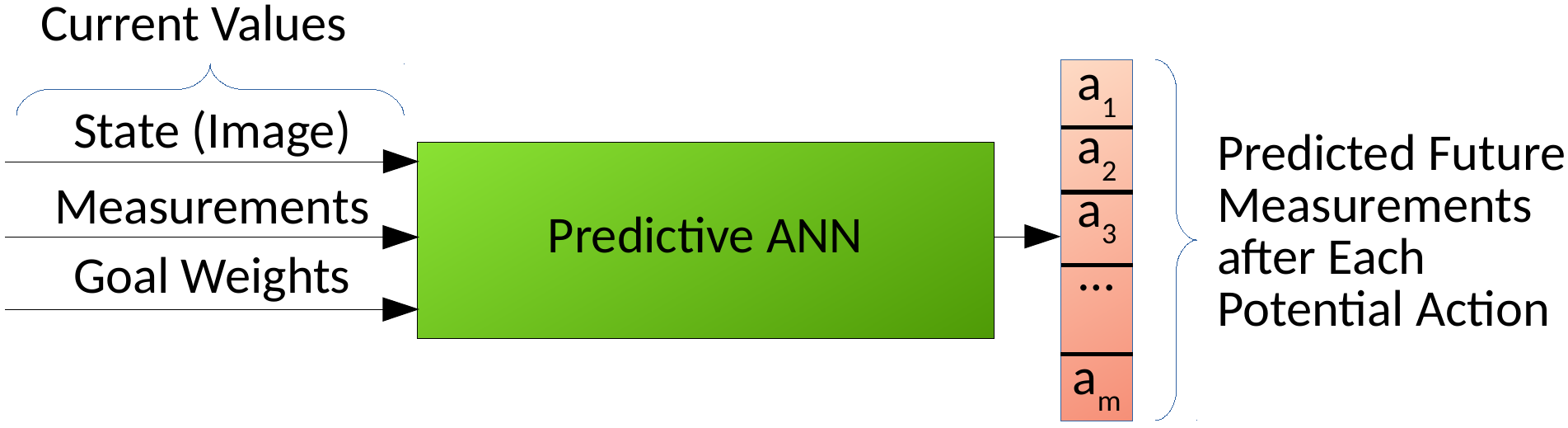}}%
	
	\caption{\textbf{Combining adaptive goals with prediction-based action 
			selection.} Top: Our proposed Goal-ANN produces a weighting of the 
		agent's 
		goals adapted to the current situation (indicated by current 
		measurements). Bottom: The resulting weights are given to the 
		predictive 
		ANN, together with the current state and current measurements, 
		resulting in 
		a vector of predicted future values for the measurements.}
	\label{fig:concept}
\end{figure}

In this paper, we explore the potential for \emph{automatically adapting} 
such goal vectors to a given scenario (Figure~\ref{fig:concept}). We show that 
this allows the reuse 
of a 
learned predictive model in new situations, where the best strategy has 
changed. The automatically adapting goal vectors, together with the 
already learned predictive model, can quickly generate behaviors for new 
scenarios where the underlying rules are the same (allowing reuse of 
learned predictions), but the goals are different. We further show that 
our adaptive goal vectors can adapt agent strategies on-line, responding 
to changes as they occur.



\section{Background}


\subsection{Model-based reinforcement learning}


The problems we target in this paper are reinforcement learning (RL) problems, 
meaning an agent is tasked with learning how to solve some problem without any 
explicit guidance -- relying instead on infrequent feedback in the form of 
rewards and punishment from the environment. Reinforcement learning can be 
divided into two high-level categories: Model-based and model-free RL. 
Model-free RL means we try to solve the problem without forming an explicit 
model of the environment, relying instead on learning a mapping from 
observations to values or actions. Many of the recent 
successes in deep RL have been in this category, including deep Q 
learning~\cite{Mnih2015}, which was the first example of power of deep RL for 
playing computer games, and the deep deterministic policy gradient algorithm 
(DDPG) which demonstrated that deep neural networks can also learn to solve 
continuous control problems~\cite{Bengio2009}.

Model-based RL attempts to solve two key challenges with model-free approaches: 
1) They require enormous amounts of training data, and 2) there is no 
straightforward way to transfer a learned policy to a new task in the same 
environment~\cite{NIPS2017_7152}. To do this, model-based RL takes the approach 
of first learning a predictive model of the environment, before using this 
model to make a plan that solves the problem. An internal predictive model 
facilitates transfer of knowledge to new tasks in 
the same environment: Once the predictions are learned, they can be used for 
planning ahead to solve many different tasks.

Despite these advantages of model-based algorithms, model-free RL has so far 
been most successful for complex environments. A key reason for this is that 
model based RL is likely to produce very bad policies if the learned predictive 
model is imperfect, which it will be for most complex 
environments~\cite{NIPS2017_7152}. Recently, algorithms have been developed 
which address this problem, for instance by learning to interpret imperfect 
predictions~\cite{NIPS2017_7152}, applying new video prediction 
techniques~\cite{NIPS2018_7512} and dynamics models~\cite{Hafner2018}, or 
periodically restarting prediction 
sequences, reducing the effect of accuracy degrading for long-term 
prediction~\cite{kaiser2019model}.

Another way to mitigate the problem of having an imperfect predictive model is 
to keep the prediction task as simple as possible. While the methods above 
generally try to predict the entire future state (more specifically, entire 
frames of input in the example of video games), one could predict more 
accurately by focusing on predicting exactly the values needed to guide action 
making. Dosovitskiy and Koltun~\cite{Dosovitskiy2017} presented impressive 
results on the 
VizDoom RL environment by teaching agents to predict just a few key 
measurements, and combining this with a \emph{goal vector} that defines a 
weighting among the measurements, effectively defining which type of future we 
most wish to observe. Here, we propose to combine this method with adaptive 
goal vectors, which has the potential to allow reuse of a predictive model 
in new scenarios where the goals are different.

\subsection{Combining deep learning and neuroevolution}

Our proposed method combines two neural networks (Figure~\ref{fig:concept}), 
which are trained in different ways. The Prediction ANN is trained in a 
self-supervised manner, using the difference between predicted and actual 
future states as loss function. For the 
Goal-ANN, we have no target value, since we do not know the ``correct'' goal.
We therefore optimize this networks by using neuroevolution, a technique 
employing a population of neural networks, having the ones performing their 
task 
best become further adapted and specialized to the problem~\cite{Stanley2019}. 

A few other papers have recently combined deep learning with 
neuroevolution, aiming to get the benefits of both. Neuroevolution (NE) 
requires far more computation to solve problems than backpropagation-based deep 
learning. On the other hand, NE does not rely on a differentiable architecture, 
and works well in problems with sparse rewards, which are a challenge for most 
deep reinforcement learning algorithms~\cite{Poulsen2017}. A promising way to 
combine NE and DL is therefore to let the deep learning do the ``heavy 
lifting'', for instance learning to make predictions or recognize objects based 
on a large number of examples, and train a small action-selection component 
using NE with the pre-trained deep neural network as a back-end. This approach 
was taken by~\cite{NIPS2018_7512}, who trained a self-supervised deep neural 
network to predict the future, and then evolved a much smaller network for 
selecting actions based on internal states in the predictive network. 
A similar idea is to train a convolutional neural network to 
translate raw pixels to compact feature 
representations, before training a smaller evolving ANN to use the learned 
features to choose the right actions in an RL problem. This has been done 
successfully for simulated car racing~\cite{koutnik2014evolving}, for 
learning to aim and shoot in a video game~\cite{Poulsen2017} and for a 
health-gathering VizDoom scenario~\cite{Alvernaz}.

Similarly to the work described above, our method lets the DL component do the 
data-intensive job, which is to learn to predict the consequences of 
actions from a large number of examples. Unlike the methods above, we only 
require our 
deep network to predict a few key future measurements, greatly simplifying 
its task, and potentially reducing the impact of errors due to faulty 
predictions.





\section{Methods}

\subsection{Learning to act by predicting the future}


The algorithm we propose in this paper is an extension of ``Direct Future 
Prediction'' (DFP) from the paper \emph{Learning to act by predicting the 
future}~\cite{Dosovitskiy2017}. DFP is based on the idea of transforming an RL 
problem into a self-supervised 
learning problem of predicting parts of future states and selecting actions 
that produce those 
states that align best with an agent's goals. A deep 
neural 
network is trained to predict the 
future value of a set of different measurements, $\vec{m}$, given the current 
state and action. $\vec{m}$ should contain metrics that give an 
indication of how successfully an agent is solving its task. The scenario 
explored here and in~\cite{Dosovitskiy2017} is a video game where an agent 
attempts to survive and kill monsters. $\vec{m}$ therefore contains 
measurements of 
an agent's ammunition, health and the number of enemies killed, which are all 
interesting indicators of an agent's performance.

If an agent can learn a model of how its actions affect future measurements, 
selecting the best action is reduced to a problem of finding which of the 
potential future $\vec{m}$-vectors correspond best with the agent's goals.
~\cite{Dosovitskiy2017} solve this by formulating a goal-vector 
$\vec{g}$, where each element indicates how much we care about the 
corresponding measurement in $\vec{m}$. For instance, given that we use 
ammunition, health and enemies killed as measurements, the goal-vector 
[0.5, 1, -1] would indicate that we are interested in collecting ammunition, 
twice as interested in collecting health, and interested in \emph{avoiding} 
attacking enemies (due to the negative value for that goal).

Training this network to predict the effect of one's actions can now be done by 
collecting many episodes of gameplay, storing at each timestep $t$ current 
measurements ($\vec{m_t}$), 
sensory input ($\vec{s_t}$ -- frames of images from gameplay in this case), 
the action taken ($a_t$) and the goal vector ($\vec{g_t}$ -- this is not 
changed during an episode while training). A sample of ($\vec{s_t}$, 
$\vec{m_t}$, $a_t$, $\vec{g_t}$) would now be the training input ($\vec{I}$), 
while 
the 
target output value ($\vec{O}$) would be the change in the measurements at 
selected
future 
timesteps ($\vec{m_{t+\tau_1}}$, ..., $\vec{m_{t+\tau_n}}$). The network is 
trained using backpropagation with the loss function depending on the 
difference between its predicted change in future measurements 
($\vec{\hat{O}}$) and 
the actual observed values ($\vec{O}$).

\subsection{Adaptive goals guiding action selection}

The contribution of this paper is a technique for adapting the goals 
$(\vec{g})$ of 
DFP-agents, which enables them to 1) transfer predictive models to new tasks 
and 2) adjust goals on-line according to current measurements. To do so, we 
take a pre-trained predictive model from the DFP-algorithm, and extend it with 
an additional neural network that suggests the best goal-vector given the 
current measurements (Figure~\ref{fig:concept}).

This Goal-ANN is trained using neuroevolution~\cite{Stanley2019}: A population 
of neural networks compete for their ability to produce relevant goals. The 
best networks are randomly changed, by adding or removing nodes and 
connections, 
while worse networks are discarded. To select between networks, they are each 
assigned a \emph{fitness score} reflecting how well they perform their task. In 
our setup, we calculate the fitness by inserting the agent governed by the 
evaluated Goal-ANN and the (pre-trained) Predictive ANN into the game scenario, 
and measuring how well it performs its task (which varies slightly between our 
different experiments -- see Section \ref{sec:results}). To counter effects of 
randomness, each evaluation tests the agent 8 times and calculates the average 
performance, 

The evolving networks are 
relatively small feed-forward ANNs where the connectivity and number of neurons 
are being optimized. The inputs are the 3 measurements ($\vec{m_t}$), and the 
output is the goal vector 
($\vec{g_t}$) with the 3 corresponding goal weights. The inputs give the 
current 
value for the amount of ammunition, health and number of enemy kills (in that 
order), and the outputs represent the weight of the ammunition-objective, the 
health-objective, and the killing monsters-objective, all in the range [-1,1].

Our setup uses the popular neuroevolution algorithm NEAT~\cite{Stanley2002}, 
more specifically the NEAT-Python 
implementation\footnote{\url{https://neat-python.readthedocs.io/}}, 
with the parameters shown in Table~\ref{table:params}. With these parameters, 
each run of the algorithm results in 5,000 evaluations, corresponding to 5,000 
1-minute episodes of game play. The original DFP-algorithm was trained for 
around 95,000 episodes, demonstrating that our proposed goal-adaptation is an 
order of 
magnitude faster than training new strategies from scratch. 

\begin{table}
\caption{Parameters for Python-NEAT}
\subfloat[Mutation Rates]{	\begin{tabular}[t]{ll}
		\hline
		Parameter &  Value\\
		\hline
		Add Connections & 0.15\\
		Delete Connection & 0.1\\
		Add Node & 0.15\\
		Delete Node & 0.1\\		
		Weight Mutation & 0.8\\
		Weight Replacement & 0.02\\
		\hline
\end{tabular}}
\hspace{1cm}
\subfloat[Other parameters]{	\begin{tabular}[t]{lp{3cm}}
		\hline
		Parameter &  Value\\
		\hline
		Population Size & 50\\
		Generations & 100\\
		ANN Inputs / Outputs & 3 / 3\\
		Weights Range & [-30, 30]\\
		Activation Function & Clamped linear response in range [-1,1]\\
		\hline
\end{tabular}}
\label{table:params}
\end{table}

\subsection{Training Scenario}


The scenario we use for training and testing the Goal-ANN is from the 
original paper suggesting DFP~\cite{Dosovitskiy2017}. The scenario is based on 
the 
VizDoom platform~\cite{Kempka2017}, a popular platform for developing and 
testing reinforcement learning algorithms. VizDoom is based on the first-person 
shooter (FPS) Doom, and offers the potential to train agents to handle complex 
3D environments directly from pixel inputs. VizDoom scenarios can be used to 
train and test skills such as understanding one's surroundings, navigation, 
exploration and dealing with opponents/enemies. The violent nature of the game 
is a concern, and we are eager to test our technique on more peaceful scenarios 
soon. However, to test extending the exact model trained 
in~\cite{Dosovitskiy2017}, we needed to reuse one of their scenarios.

The specific VizDoom scenario we use here is the one titled ``D3-Battle'' 
in~\cite{Dosovitskiy2017}. This is a challenging scenario where the agent is 
under attack by alien monsters inside a maze, and has to try to kill as many of 
the monsters as possible. To aid the agent, ammunition and health kits are 
scattered around the maze. The agent is provided with (and learns to predict) 
three measurements: Its current amount of ammunition and health, as well as how 
many enemies it has killed. We use a trained predictive model from the original 
paper, which was trained in a ``goal-agnostic'' manner, that is, the goal 
vector $\vec{g}$ was randomized between episodes (each value uniformly sampled 
from the interval [-1, 1]). Such goal-agnostic training was found to generalize 
better to new tasks than fixed-goal predictors~\cite{Dosovitskiy2017}, and we 
therefore focus on this model for our study.





\section{Results}
\label{sec:results}


We want to explore two possible advantages of evolved goal weights: 1) 
Their ability to adapt an existing predictive model to new scenarios where 
strategies need to be different, and 2) Their ability to produce 
context-dependent goals, that is, goals that vary depending on 
current measurements.

To do so, we compare three main techniques: 1) Acting by following the same 
static goal-vector applied in~\cite{Dosovitskiy2017}, 2) Acting by 
following a simple rule for adapting the goal to the current situation and 3) 
Acting by following the evolved Goal-ANN. The static goal vector 
from~\cite{Dosovitskiy2017} is [0.5, 0.5, 1.0] where the 
three numbers represent the importance of the ammunition-, health- and enemy 
kills-objectives, respectively. This goal captures the intuition that the aim 
of the game is to kill as many monsters as possible, but collecting some 
ammunition and health is a good side-goal to help the agent in its primary 
mission. The simple rule to improve this static goal (called ``Hardcoded'' in 
plots below) is to switch to the goal [0, 1.0, -1.0] whenever the agent's 
health is below 50\%, which adds the intuition that we should stop attacking 
and focus on gathering health when injured.

In all plots below, fitness values are averaged over 20 independent evaluations 
of 
each agent.


\subsection{The original scenario}

\begin{wrapfigure}{R}{0.55\textwidth}
	\centering
	\includegraphics[width=0.5\textwidth]{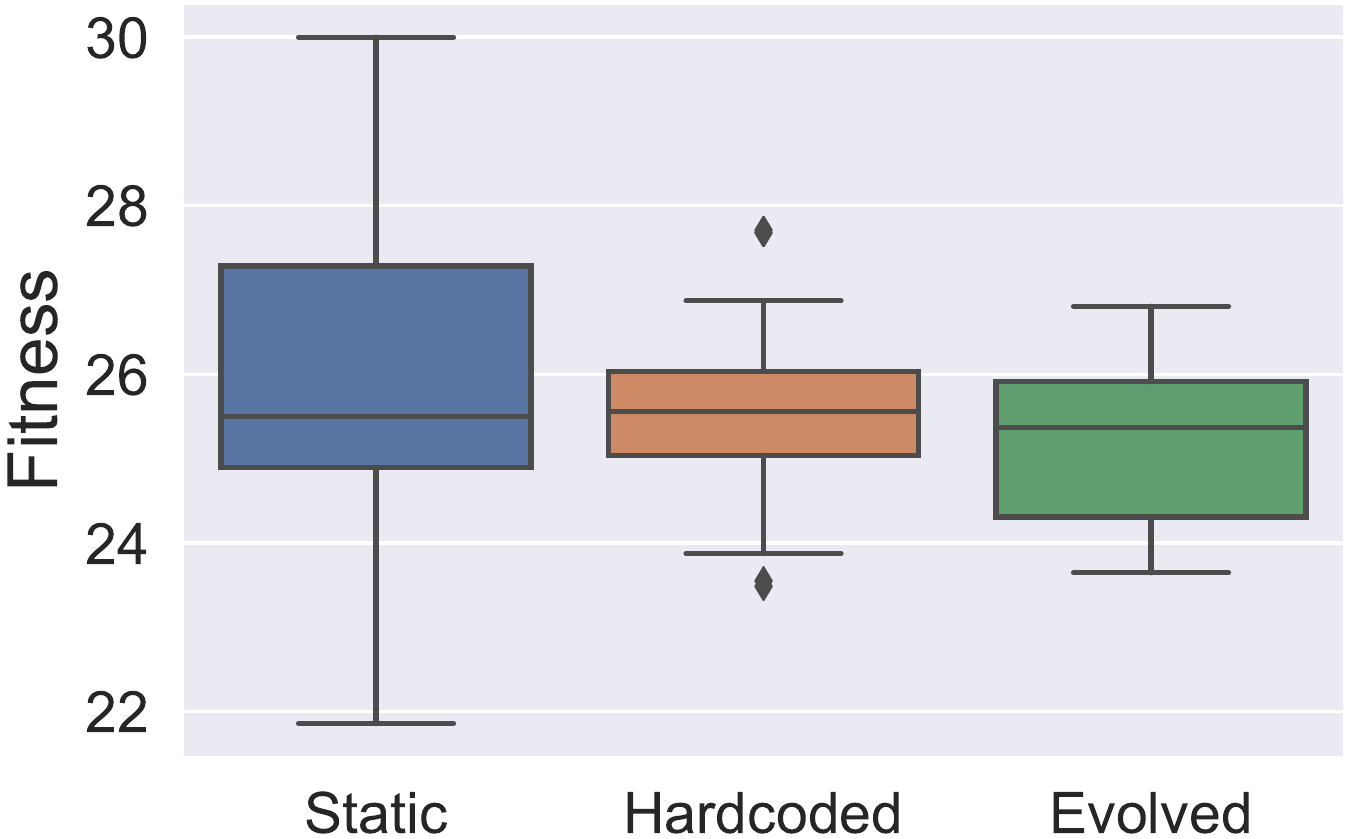}
	
	\caption{\textbf{The original scenario.} No significant differences in 
		fitness values between using the original goal (Static), a simple 
		adaptive 
		goal (Hardcoded) and the evolved Goal-ANN.}
	\label{fig:results_original}
	
	\vspace{-30pt}
\end{wrapfigure}

As mentioned above, we test our method on a VizDoom scenario developed 
by~\cite{Dosovitskiy2017}, using their pre-trained agent together with our 
evolved adaptive goals. The scenario consists of a maze populated by 
monsters and the agent. The goal of the trained agent is to kill as many 
monsters as 
possible, and to do so, it can benefit from collecting ammunition and health 
packages along the way. In this original scenario, the final reward (also 
referred to as Fitness below, as 
is common in Evolutionary Algorithms) is based only on the number of monsters 
killed.


Figure~\ref{fig:results_original} shows the average reward (number of monsters 
killed in one minute) of the three 
compared techniques on the original scenario. There are no significant 
differences between the compared techniques. In other words, there is no 
advantage to adapting the goals in this scenario.

Analyzing the agents' gameplay in this scenario reveals why this is the case: 
Agents are very fast, killing monsters without taking much damage, which makes 
the default goal 
of aggressively attacking monsters while picking up any health or ammo ahead a 
very good choice. In other words, there is no conflict among the goals and 
no reason why a different choice of goal weights should improve 
performance.


\subsection{A hard scenario}

To test the ability of the adaptive goals to solve different scenarios, 
and the potential for adjusting goals during gameplay, we set up a much 
harder scenario. We introduce the following difficulties that 
force the agent to 
sometimes change its strategy between defensive and aggressive modes:

\begin{itemize}
	\item Monsters are tougher (they have twice as much health as before).
	\item Player is weaker (it begins the game with only 10\% health)
	\item A fitness penalty of 100 is given for dying (versus 0 before). 
	\item Player starts with 0 initial ammunition (versus 20 bullets before).
\end{itemize}

\begin{figure}
	\centering
	\subfloat[Fitness]{\includegraphics[width=0.48\textwidth]{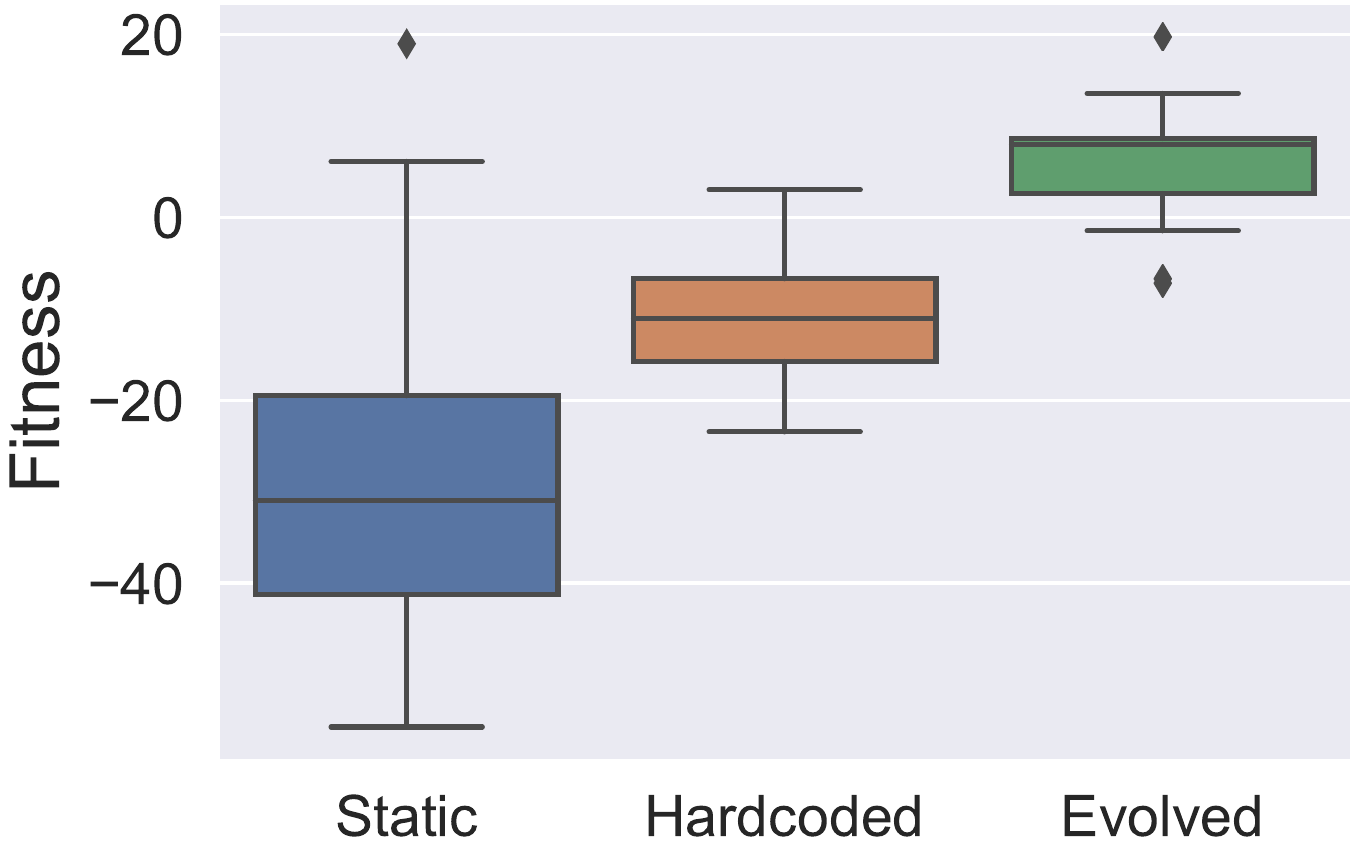}\label{fig:hard_fitness}}%
	\hfill
	\subfloat[Evolving 
	Goals]{\includegraphics[width=0.465\textwidth]{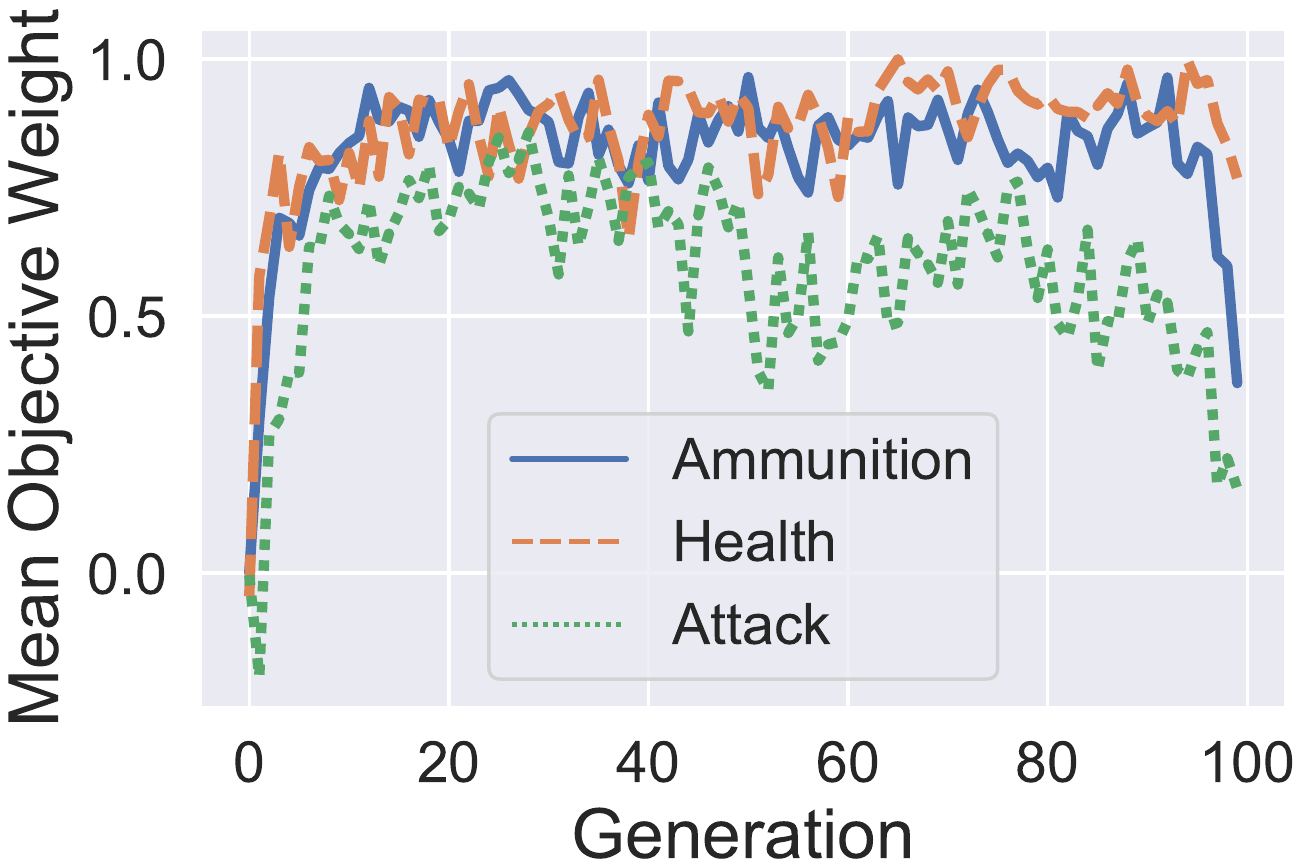}\label{fig:hard_obj_generations}}
	\caption{\textbf{The hard scenario}. Left: The evolved adaptive Goal-ANN 
		significantly outperforms both the standard (static) and hardcoded 
		(dynamic) 
		goals. 
		Right: The mean population value for each goal through evolution. The 
		evolved strategies tend to focus more on health and less on attacking.}
	\label{fig:results_hard}
\end{figure}



Since the monsters are now stronger and the player weaker, the default static 
goals result in a strategy that is too aggressive and frequently results 
in the player dying. The hardcoded strategy performs better, since it balances 
aggression and defense. The evolved strategy significantly outperforms both the 
others ($p<0.05$), finding an even better balance between the three objectives 
(Figure~\ref{fig:hard_fitness}).

We can see what the evolved strategy has learned by plotting the output of the 
Goal-ANN. In Figure~\ref{fig:hard_obj_generations} we plot the output of 
the Goal-ANN per generation of evolution, averaged both over the entire 
population and all timesteps in each individual's life. We see that the evolved 
strategies gradually move towards a focus on the health-objective, while 
reducing focus on attacking enemies.

\begin{wrapfigure}{R}{0.55\textwidth}
	\centering
	\includegraphics[width=0.5\textwidth]{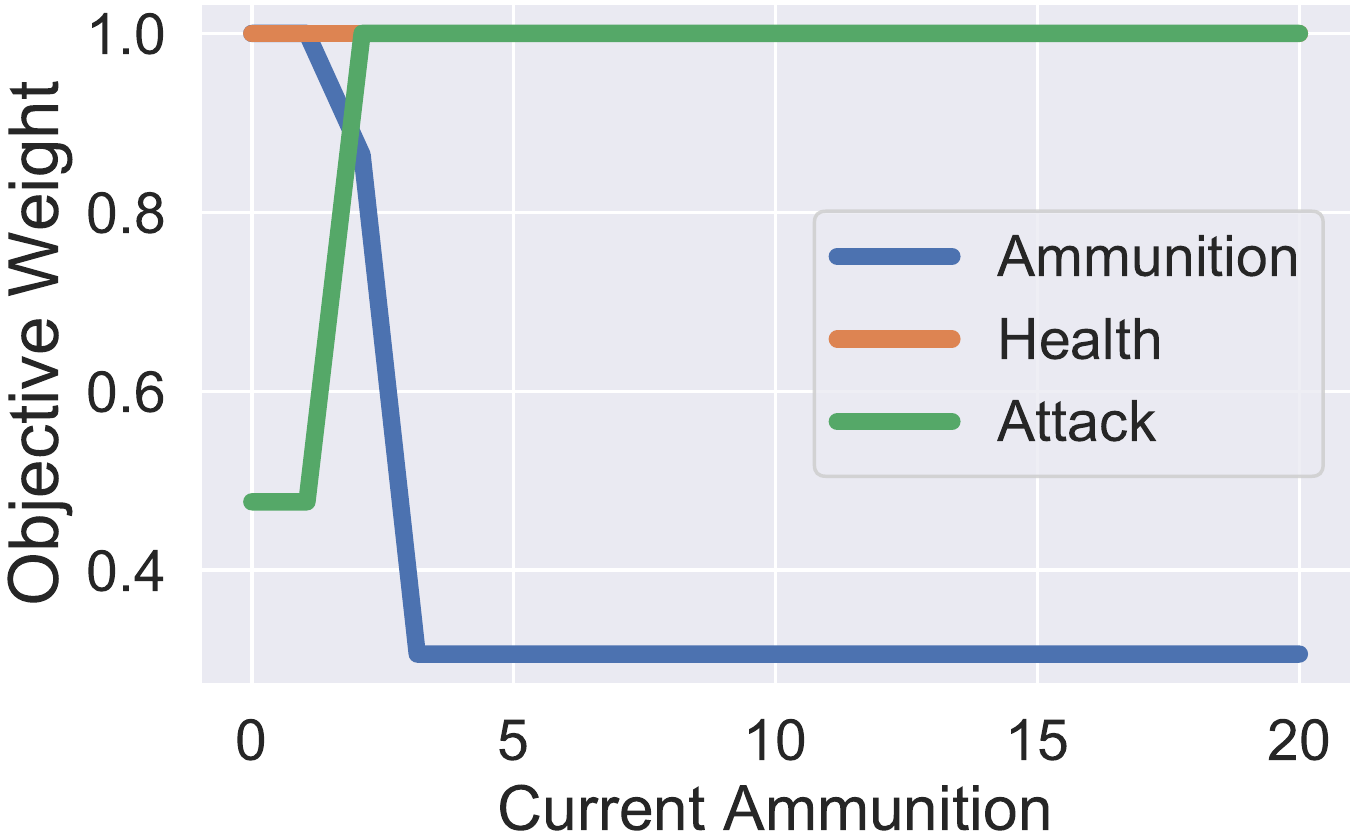}
	\caption{\textbf{The hard scenario.} The strategy discovered by evolution 
		is to focus on gathering 
		ammunition until some has been collected, after which the agent 
		switches to 
		attacking. Simultaneously, the agent focuses on collecting 
		health.}
	\label{fig:hard_behavior_plot}
\end{wrapfigure}

Figure~\ref{fig:hard_obj_generations} summarizes strategies over 
complete game episodes. However, the Goal-ANN has the potential to 
produce strategies that vary through a game, depending on the current 
measurements. To investigate if this is taking place, we performed a sweep over 
sensible values for all three measurements, passed these values through the 
best 
evolved Goal-ANN and measured the resulting output goals. The results show 
that changing the current health or the number of kills has no effect on the 
produced goals. However, changing the amount of ammunition does affect 
the output goals, as plotted in Figure~\ref{fig:hard_behavior_plot}. We 
see that evolution has discovered the strategy of focusing less on attacking 
and more on gathering ammunition when the current ammunition level is 
low\footnote{\url{https://youtu.be/NCzrO5KHMXQ} shows an 
	agent playing according to this strategy.}.



%
%


\subsection{The no-ammunition scenario}

To test the potential for the Goal-ANN to adapt strategies to scenarios 
where rules are very different, we set up a scenario with no ammunition 
available to the agent. The scenario is otherwise identical to the hard 
scenario above. This change turns the game into a defensive exercise, where 
staying away from enemies and gathering health is the best strategy. Since the 
aggressive default goal now results in a very bad strategy, we here test an 
additional \emph{defensive} goal, which has the maximum negative weight on the 
attack-objective, and maximum positive weights on the two others 
($\vec{g}=[1,1,-1]$).

\begin{figure}
	\centering
	\subfloat[Fitness]{\includegraphics[width=0.48\textwidth]{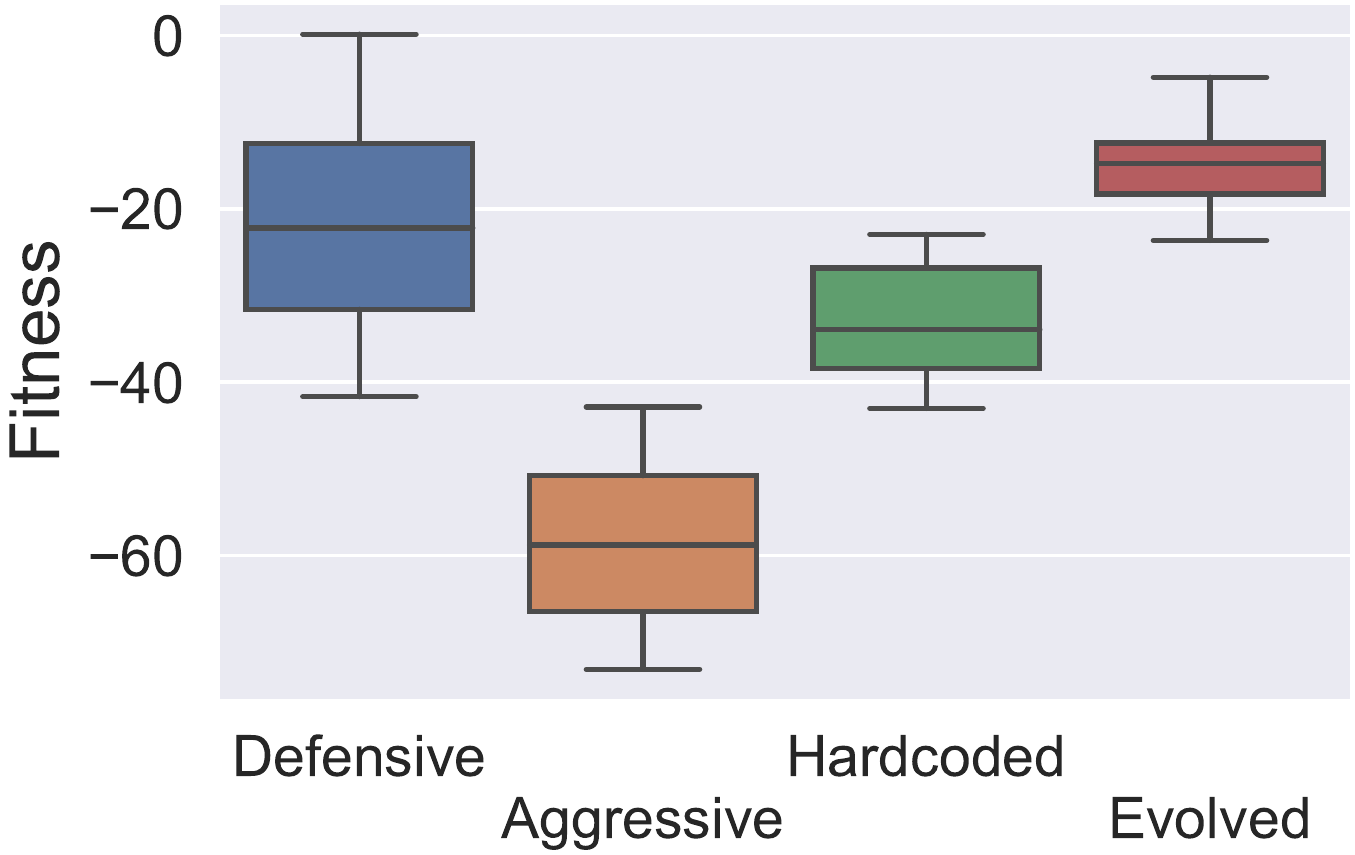}\label{fig:no_ammo_fitness}}%
	\hfill
	\subfloat[Evolving 
	Goals]{\includegraphics[width=0.48\textwidth]{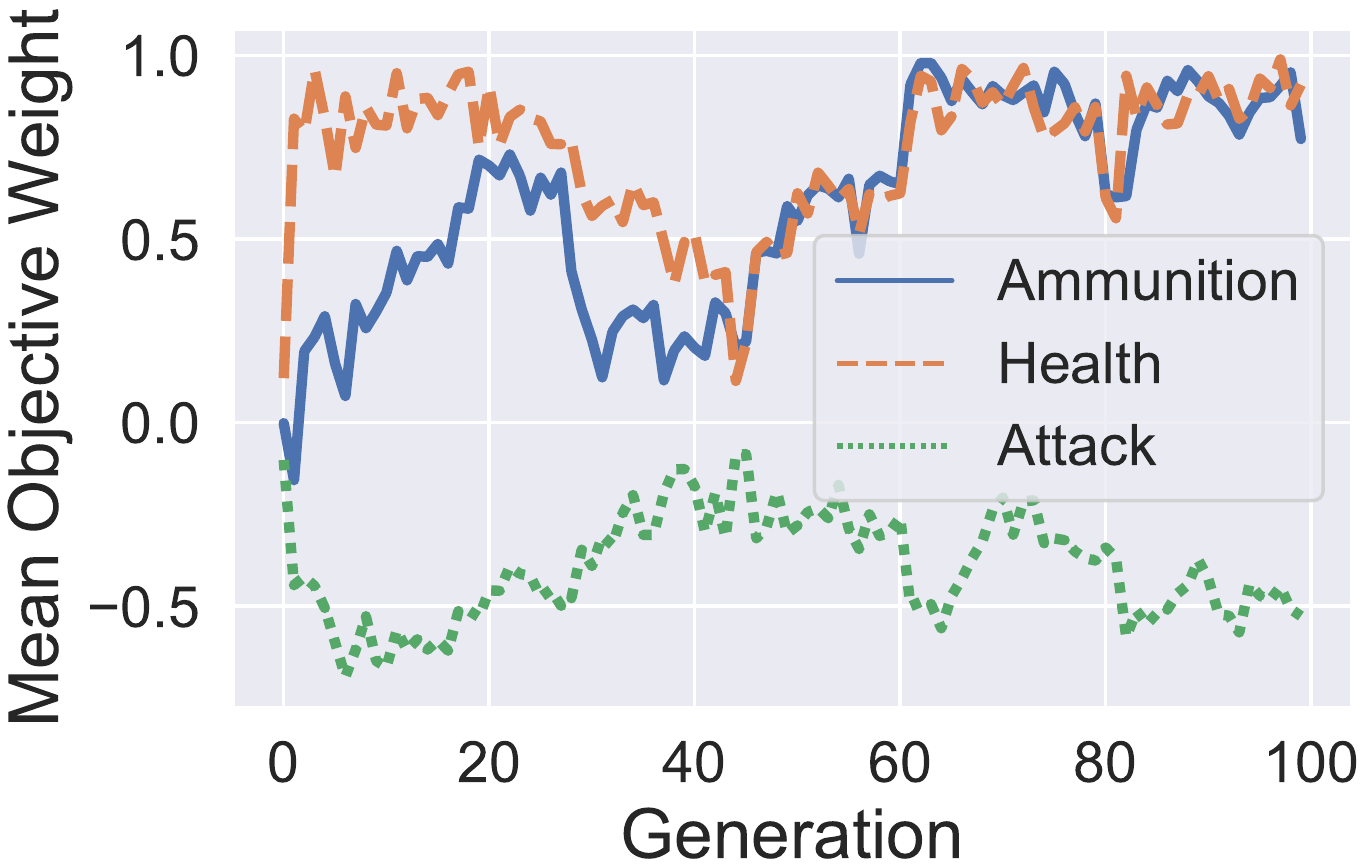}\label{fig:no_ammo_generations}}
	\caption{\textbf{The scenario with no ammunition.} Left: Evolved adaptive 
		goals generated by the Goal-ANN
		significantly outperform both the static (Aggressive and Defensive) and 
		Hardcoded 
		goals. 
		Right: The mean population value for each goal through evolution. 
		The 
		evolved strategies quickly learn to not focus on attacking.}
	\label{fig:no_ammo_scenario}
\end{figure}

As expected, the default aggressive strategy performs very badly in this 
scenario, while the hardcoded and the new defensive strategy do better 
(Figure~\ref{fig:no_ammo_fitness}). The evolved strategy significantly 
outperforms all others, with $p<0.01$ according to the Mann-Whitney U test, for 
all pairwise comparisons. We were initially surprised to see the 
evolved strategy outperform the defensive strategy, since we expected its full 
focus on avoiding confrontation to be optimal. We therefore took a closer look 
at the evolved strategy\footnote{\url{https://youtu.be/6pTnkCGV6NI} shows an 
agent playing according to this strategy.}.

\begin{wrapfigure}{R}{0.55\textwidth}
	\centering
	\includegraphics[width=0.5\textwidth]{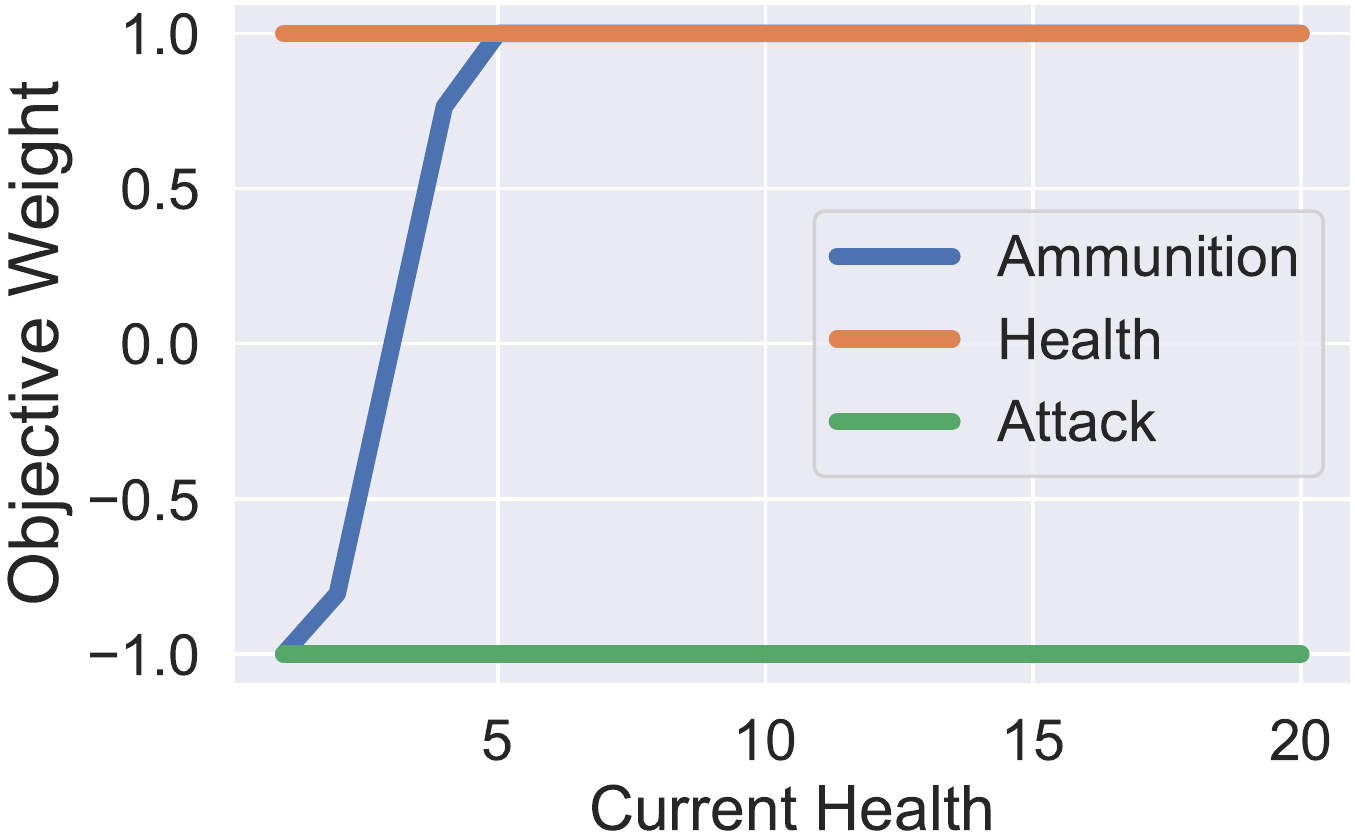}
	\caption{\textbf{The scenario with no ammunition.} The strategy discovered 
	by evolution is to focus on only on health 
	if it is low, but add focus on gathering ammunition otherwise.}
	\label{fig:no_ammo_strategy}
\end{wrapfigure}

As expected, we found the evolving strategy to give a strong negative weight to 
the attack-objective, and strong positive weights to the two others 
(Figure~\ref{fig:no_ammo_generations}). Repeating the sweep across 
measurements, we found that the current health measure is the only value that 
leads to different goals when changed. The evolved strategy is to focus 
exclusively on collecting health when the measure is critically low, and 
otherwise to also include a drive for collecting ammunition, in both cases 
avoiding attacking as strongly as possible (Figure~\ref{fig:no_ammo_strategy}). 
It may seem surprising that the ammunition objective is valuable at all here, 
since there is no ammunition in 
the environment. We hypothesize that a high value for this objective can be 
valuable, since it can encourage the agent to keep moving, heading towards 
objects with a small resemblance of ammunition, thus staying away from danger.

\section{Conclusion}

We proposed extending an existing technique for selecting actions based on 
predictions of the consequences of those actions with an adaptive 
goal-producing neural network. This Goal-ANN can modify the behavior resulting 
from predictions of the future, by changing the agent's preference among 
different future outcomes.

We demonstrated that the Goal-ANN allows the transfer of a predictive model to 
scenarios where the optimal behavior is different, due to different amounts of 
resources in the environment. Such knowledge transfer has the potential to 
greatly improve the efficiency of Reinforcement Learning methods, since they 
allow training a model on one task, and adapting it rapidly to other, related 
problems.

We also showed that the Goal-ANN is capable of learning \emph{adaptive} 
strategies, where goals change on-line, depending on the current state of the 
environment. This is a very valuable property, since most complex problems 
require one to modify one's strategy depending on the current context.

A final valuable feature of the Goal-ANN is its \emph{interpretability}: We can 
easily probe the rules it has learned by sweeping across input measurements and 
observing the resulting output goals. We demonstrated that in our target 
scenarios, such an analysis allows us to verify that the learned goal-switching 
behavior is sensible.

This has been an initial study of the potential for combining self-adapting 
goals with a predictive deep neural network. There are many future questions 
here that we aim to address, including tests on more scenarios, comparison with 
other techniques for transfer learning and testing if the Goal-ANN could 
perform even better if given more information about the current state of the 
world.




\section{Acknowledgments}

This work is supported by The Research Council of Norway as part of the 
Engineering Predictability with Embodied Cognition (EPEC) project $\#$240862, 
and the Centres of Excellence scheme, project $\#$262762.
%
%
%
\bibliographystyle{splncs04}
\bibliography{bibliography_without_urls}

\end{document}